\documentclass{article}

\usepackage{microtype}
\usepackage{amsmath}
\usepackage{graphicx}
\usepackage{amsthm}
\usepackage{multirow}
\usepackage{subfigure}
\usepackage{booktabs}
\usepackage{amssymb}
\usepackage{url}
\usepackage{hyperref}
\makeatletter
\g@addto@macro{\UrlBreaks}{\UrlOrds}
\makeatother
\usepackage[accepted]{icml2019}

\icmltitlerunning{Deep Residual Output Layers for Neural Language Generation}
\begin{document}

\twocolumn[
\icmltitle{Deep Residual Output Layers for Neural Language Generation}

% It is OKAY to include author information, even for blind
% submissions: the style file will automatically remove it for you
% unless you've provided the [accepted] option to the icml2019
% package.

% List of affiliations: The first argument should be a (short)
% identifier you will use later to specify author affiliations
% Academic affiliations should list Department, University, City, Region, Country
% Industry affiliations should list Company, City, Region, Country

% You can specify symbols, otherwise they are numbered in order.
% Ideally, you should not use this facility. Affiliations will be numbered
% in order of appearance and this is the preferred way.
\icmlsetsymbol{equal}{*}

\begin{icmlauthorlist}
\icmlauthor{Nikolaos Pappas}{to}
\icmlauthor{James Henderson}{to} 
\end{icmlauthorlist}

\icmlaffiliation{to}{Idiap Research Institute, Martigny, Switzerland}
\icmlcorrespondingauthor{Nikolaos Pappas}{nikolaos.pappas@idiap.ch}

% You may provide any keywords that you
% find helpful for describing your paper; these are used to populate
% the "keywords" metadata in the PDF but will not be shown in the document
\icmlkeywords{Machine Learning, ICML}

\vskip 0.3in
]

% this must go after the closing bracket ] following \twocolumn[ ...

% This command actually creates the footnote in the first column
% listing the affiliations and the copyright notice.
% The command takes one argument, which is text to display at the start of the footnote.
% The \icmlEqualContribution command is standard text for equal contribution.
% Remove it (just {}) if you do not need this facility.

%\printAffiliationsAndNotice{}  % leave blank if no need to mention equal contribution
%\printAffiliationsAndNotice{\icmlEqualContribution} % otherwise use the standard text.
\printAffiliationsAndNotice{} 

\begin{abstract}

Many tasks, including language generation, benefit from learning the structure of the output space, particularly when the space of output labels is large and the data is sparse. 
State-of-the-art neural language models indirectly capture the output space structure in their classifier weights since they lack parameter sharing across output labels. Learning shared output label mappings helps, but existing methods have limited expressivity and are prone to overfitting.
In this paper, we investigate the usefulness of more powerful shared mappings for output labels, and propose a deep residual output mapping with dropout between layers to better capture the structure of the output space and avoid overfitting. 
Evaluations on three language generation tasks show that our output label mapping can match or improve state-of-the-art recurrent and self-attention architectures, and 
suggest that the classifier does not necessarily need to be high-rank to better model natural language if it is better at capturing the structure of the output space.

\end{abstract}
 
\section{Introduction}

Learning the structure of the output space benefits a wide variety of tasks, such as object recognition and novelty detection in images \cite{Weston2011,Socher13,frome13,ZhangGS16,Chen_2018_CVPR}, zero-shot prediction in texts \cite{dauphin14,yazhen15,nam16,rios-kavuluru-2018-shot}, and structured prediction in either images or text \cite{NIPS2014_5323,dyer2015,Belanger2016,NIPS2018_7869}.  When the space of output labels is large or their data is sparse, treating labels as independent classes makes learning difficult, because identifying one label is not helped by data for other labels.  This problem can be addressed by learning output label embeddings to capture the similarity structure of the output label space, so that data for similar labels can help classification, even to the extent of enabling few-shot or even zero-shot classification. This approach has been particularly successful in natural language generation tasks, where word embeddings give a useful similarity structure for next-word-prediction in tasks such as machine translation \citep{NIPS2017_7181} and language modeling \citep{merity2017regularizing}.

Existing neural language models typically use a log-linear classifier to predict words \cite{NIPS2017_7181,chen2018}. We can view the output label weights as a word embedding, and the input encoder as mapping the context to a vector in the same embedding space.  Then the similarity between these two embeddings in this joint input-label space is measured with a dot product followed by the softmax function. We will refer to this part as the classifier, distinct from the input encoder which only depends on the input and the label encoder which only depends on the label. To improve performance and reduce model size, sometimes the output label weights are tied to the input word embedding vectors  \cite{inan2016tying,press17}, but there is no parameter sharing taking place across different words, which limits the effective transfer between them.
 
Recent work has shown improvements over specific vanilla recurrent architectures by sharing parameters across outputs through a bilinear mapping on neural language modeling \cite{gulordava18} or a dual nonlinear mapping on neural machine translation \cite{pappas18}, which can make the classifier more powerful. However, the shallow modeling constraints and the lack of regularization capabilities limit their applicability on arbitrary tasks and model architectures. Orthogonal to these studies, \citet{mos2018} achieved state-of-the-art improvements on language modeling by increasing the power of the classifier using a mixture of softmax functions, albeit at the expense of computational efficiency. A natural question arises of whether one can make the classifier more powerful by simply increasing the power of the label mapping while using a single softmax function without modifying its dimensionality or rank. 

In this paper, we attempt to answer this question by investigating alternative neural architectures for learning the embedding of an output label in the joint input-label space which address the aforementioned limitations. In particular, we propose a deep residual nonlinear output mapping from word embeddings to the joint input-output space, which better captures the output structure while it avoids overfitting with two different dropout strategies between layers, and preserves useful information with residual connections to the word embeddings and, optionally, to the outputs of previous layers.\footnote{Our code and settings are available at \url{http://github.com/idiap/drill}.} For the rest of the model, we keep the same input encoder architecture and still use the dot product and softmax function for output label prediction.  
 
We demonstrate on language modeling and machine translation that we can match or improve state-of-the-art recurrent and self-attention architectures by simply increasing the power of the output mapping, while using a single softmax operation and without changing the dimensionality or rank of the classifier. The results suggest that the classifier does not necessarily need to be high rank to better model language if it better captures the output space structure. Further analysis reveals the significance of different model components and  improvements on predicting low frequency words.

\section{Background: Neural Language Generation}
\label{background}
  
The output layer of neural models for language generation tasks such as language modeling \cite{bengio2003,mikolov2012context,merity2017regularizing},  machine translation \cite{Bahdanau15,luong15,TACL1081} and summarization \cite{rush15,paulus2018a}, typically consists of a linear unit with a weight matrix $\mathbf{W}\in \rm I\!R^{d_h \times |\mathcal{V}|}$ and a bias vector  $\mathbf{b}\in \rm I\!R^{|\mathcal{V}|}$ followed by a softmax activation function,  where $\mathcal{V}$ is the vocabulary.  
Thus, at a given time $t$, the output probability distribution for the current output  $\mathbf{y_t}$ conditioned on the inputs i.e.\ the previous outputs, $\mathbf{y_1^{t-1}} = (\mathbf{y_1}, \mathbf{y_2}, \cdots, \mathbf{y_{t-1}})$ with $\mathbf{y_i}\in\{0,1\}^{|\mathcal{V}|}:  \sum^{|\mathcal{V}|}_j \mathbf{y_{i}}_j = 1 \ \forall \ i \in \mathcal{N} $, is defined as:
\begin{align}
 p(\mathbf{y_t}|\mathbf{y_1^{t-1}}) & \propto  \texttt{exp}(\mathbf{W}^T\mathbf{{h}_t} + \mathbf{b}) ,
          \label{baseline_eq}
\end{align}
where $\mathbf{{h}_t}$ is the input encoder's hidden representation at time $t$ with $d_h$ dimensions. The parameterisation in Eq.~\ref{baseline_eq} makes it difficult to learn the structure of the output space or to transfer this information from one label to another because the parameters for output label $i$,  $\mathbf{W}^T_i$, are independent from the parameters for any other output label $j$, $\mathbf{W}^T_j$.

\subsection{Weight Tying} Learning the structure of the output space can be helped by learning it jointly with the structure of input word embeddings, but this still does not support the transfer of learned information across output labels.  In particular, since the output labels are words and thus the output parameters $\mathbf{W}^T$ have one row per word, it is common to tie these parameters with those of the input word embeddings $\mathbf{E} \in \rm I\!R^{|\mathcal{V}| \times d}$, by setting $\mathbf{W} = \mathbf{E}^T$ \cite{inan2016tying,press17}. 
Making this substitution in Eq~\ref{baseline_eq}, we obtain: 
\begin{align}
 p(\mathbf{y_t}|{\mathbf{y_1^{t-1}}}) 
          & \propto  \texttt{exp}(\mathbf{E} \mathbf{{h}_t} + \mathbf{b})  \label{eq:tied} 
\end{align} 
Although there is no explicit transfer across outputs, this parameterisation can implicitly learn the output structure, as can be seen if we assume an implicit factorization of the input embeddings, $\mathbf{E}\approx \mathbf{E_{l}} \mathbf{W_l}$ as in \cite{mikolov13}.

\subsection{Bilinear Mapping}  
The above bilinear form, excluding the bias, is similar to the form of joint input-output space learning models \cite{yazhen15,nam16} which have been proposed in the context of zero-shot text classification. This motivates the learning of explicit relationships across outputs and inputs through parameter sharing via $\mathbf{W_l}$ as above. By substituting this factorization in Eq~\ref{eq:tied}, we obtain: 
\begin{align}
 p(\mathbf{y_t}|{\mathbf{y_1^{t-1}}}) 
          & \propto  \texttt{exp}(\mathbf{E_{l}}\mathbf{\mathbf{W_l}}\mathbf{{h}_t} + \mathbf{b})   \label{eq:bilinear_soft}
\end{align} 
where $\mathbf{W_l} \in \rm I\!R^{d \times d_h}$ is the bilinear mapping and $\mathbf{E}, \mathbf{h_t}$ are the output embeddings and the encoded input respectively, as above. This parametrization has been previously also proposed by \citet{gulordava18} for language modeling albeit with a different motivation, namely to decouple the hidden state from the word embedding prediction.

\subsection{Dual Nonlinear Mapping} \label{joint} Another existing output layer parameterisation which explicitly learns the structure of the output is from \cite{pappas18}. Specifically, two nonlinear functions, $g_{out}(\cdot)$ and $g_{in}(\cdot)$, are introduced which aim to capture the output and context structure respectively: 
\begin{align}
\hspace{-3mm}p(\mathbf{y_t}|\mathbf{y_1^{t-1}}) & \propto  \texttt{exp}\big(    g_{out}(\mathbf{E}) g_{in}(\mathbf{h_t}) + \mathbf{b} \big), 
 \label{general_joint_form}
 \\
 & \propto  \texttt{exp}\big( \sigma(\mathbf{EU} + \mathbf{b_{u}}) \sigma( \mathbf{V h_t} + \mathbf{b_{v}}) + \mathbf{b} \big)
 \label{joint_form}
\end{align}
\noindent where $\sigma(\cdot)$ is a nonlinear activation function such as ReLU or Tanh, the matrix $\mathbf{U} \in \rm I\!R^{d \times d_j}$ and bias $\mathbf{b_{u}} \in \rm I\!R^{d_j}$ are the linear projection of the encoded outputs, and the matrix $\mathbf{V} \in \rm I\!R^{ d_j \times  d_h}$ and bias $\mathbf{b_{v}} \in \rm I\!R^{d_j}$ are the linear projection of the context, and $\mathbf{b} \in \rm I\!R^{\mathcal{V}}$ captures the biases of the target outputs in the vocabulary.

The parameterisation of Eq.~\ref{joint_form} enables learning a more rich output structure than the bilinear mapping of Eq.~\ref{eq:bilinear_soft} because it learns nonlinear relationships.  Both, however, allow for controlling the capacity of the output layer independently of the dimensionality of the context $\mathbf{h_t}$ and the word embedding $\mathbf{E}$, by increasing the breadth of the joint projection, e.g.~the dimensionality of the $\mathbf{U}$ and $\mathbf{V}$ matrices in Eq.~\ref{joint_form} above. This increased capacity can be seen in the inequalities below for the number of parameters of the output layers discussed so far, assuming a fixed $|\mathcal{V}|$, $d$, $d_{h}$:
\begin{align}
\mathcal{C}_{tied} < \mathcal{C}_{bilinear} \leq \mathcal{C}_{dual} \leq \mathcal{C}_{base},
\label{capacity_ineq}
\end{align}
where ${C}_{tied}$,  $\mathcal{C}_{base}$,  $\mathcal{C}_{bilinear}$ and $\mathcal{C}_{dual}$ respectively correspond to the number of dedicated parameters of an  output layer with (Eq.~\ref{eq:tied}) and without (Eq.~\ref{baseline_eq}) weight tying, using the bilinear mapping (Eq.~\ref{eq:bilinear_soft}) and the dual nonlinear mapping (Eq.~\ref{joint_form}) which are assumed to be nonzero except $\mathcal{C}_{tied}$. 
    
Given this analysis, we identify and aim to address the following limitations of the previously proposed output layer parameterisations for language generation: \
\begin{itemize}
    \item[(a)] \textit{Shallow modeling of the label space.} Output labels are mapped into the joint space with a single (possibly nonlinear) projection.  Its power can only be increased by increasing the dimensionality of the joint space.
    \item[(b)] \textit{Tendency to overfit}. Increasing the dimensionality of the joint space and thus the power of the output classifier can lead to undesirable effects such as overfitting in certain language generation tasks, which limits its applicability to arbitrary domains.   
\end{itemize}

\section{Deep Residual Output Layers} 
\label{model}
To address the aforementioned limitations we propose a deep residual output layer architecture for neural language generation which performs deep modeling of the structure of the output space while it preserves acquired information and avoids overfitting. Our formulation adopts the general form and the basic principles of previous output layer parametrizations which aim to capture the output structure explicitly in Section \ref{joint}, namely (i) learning rich output structure, (ii) controlling the output layer capacity independently of the dimensionality of the vocabulary, the encoder and the word embedding, and, lastly, (iii)  avoiding costly label-set-size dependent parameterisations.  

\subsection{Overview}
\begin{figure}[htp]
 	\centering
 	\hspace{5mm}\includegraphics[scale=0.52]{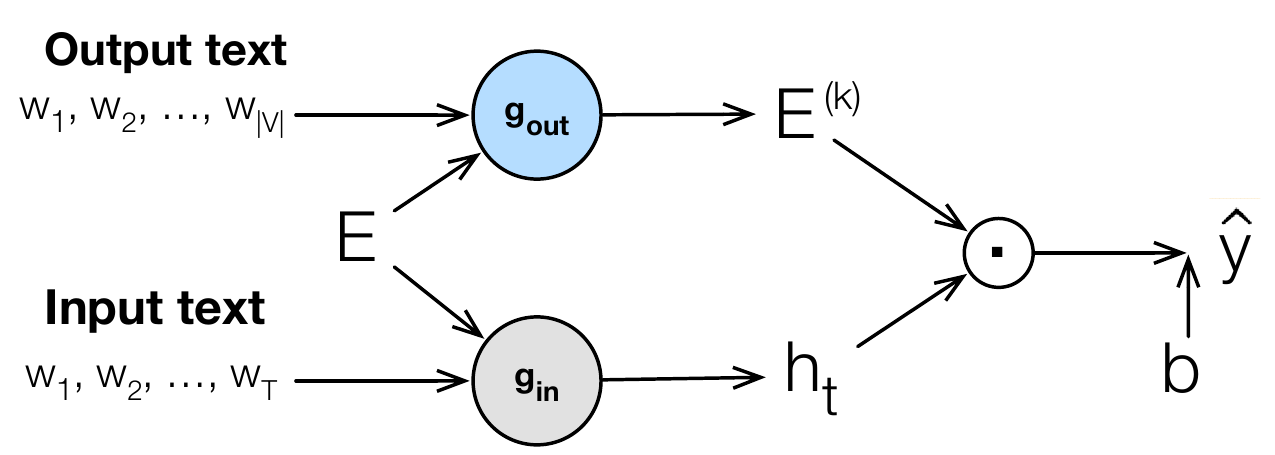}
 	\vspace{-3mm}
 	\caption{General overview of the proposed architecture.}
 	\vspace{-2mm}
 	\label{drill_schema}
\end{figure}

A general overview of the proposed architecture for neural language generation is displayed in Fig.~\ref{drill_schema}.
We base our output layer formulation starting on the general form of the dual nonlinear mapping of Eq.~\ref{general_joint_form}:
\begin{align}
    p(\bf y_t |y_1^{t-1})  & \propto \texttt{exp}\big( g_{out}(\mathbf{E}) g_{in}(\mathbf{h_t}) + \mathbf{b}\big). 
    \label{drill_form}
\end{align} 
The input network $g_{in}(\cdot)$  takes as input a sequence of words represented by their input word embeddings $\mathbf{E}$ which have been encoded in a context representation $\mathbf{h_t}$ for the given time step $t$.  The output or label network  $g_{out}(\cdot)$ takes as input the word(s) describing each possible output label and  encodes them in a label embedding $\mathbf{E^{(k)}}$ where $k$ is the depth of the label encoder network. Next, we define these two proposed networks, and then we discuss how the model is trained and how it relates to previous output layers.
 
\subsection{Label Encoder Network}
 
For language generation tasks, the output labels are each a word in the vocabulary $\mathcal{V}$.  We assume that these labels are represented with their associated word embedding, which is a row in $\mathbf{E}$.  In general, there may be additional information about each label, such as dictionary entries, cross-lingual resources, or contextual information, in which case we can add an initial encoder for these descriptions which outputs a label embedding matrix $\mathbf{E'} \in \rm I\!R^{|\mathcal{V}|\times d}$.  In this paper we make the simplifying assumption that $\mathbf{E'} = \mathbf{E}$ and leave the investigation of additional label information to future work.

\subsubsection{Learning Output Structure}
To obtain a label representation which is able to encode rich output space structure, we define the $g_{out}(\cdot)$ function to be a deep neural network with $k$ layers which takes the label embedding $E$ as input and outputs its deep label mapping at the last layer, $g_{out}(\mathbf{E}) = \mathbf{{E}^{(k)}}$,  as follows: 
\begin{align}
     \mathbf{{E}^{(k)}}    & = f^{(k)}_{out}(\mathbf{{E}^{(k-1)}}) \label{kth_proj}, 
\end{align}
\noindent where $k$ is the depth of the network and 
each function $f^{(i)}_{out}(\cdot)$ at the $i_{th}$ layer is a nonlinear projection of the following form: 
\begin{align}
    f^{(i)}_{out}(\mathbf{{E}^{(i-1)}}) = \sigma(\mathbf{E^{(i-1)}}\mathbf{U^{(i)}} + \mathbf{b^{(i)}_u}),
    \label{drill_proj}
\end{align}
where $\sigma(\cdot)$ is a nonlinear activation function such as $\texttt{ReLU}$ or $\texttt{Tanh}$, and the matrix $\mathbf{U^{(i)}} \in \rm I\!R^{d \times d_j}$ and the bias $\mathbf{b^{(i)}_{u}} \in \rm I\!R^{d_j}$ are the linear projection of the encoded outputs at the $i_{th}$ layer. Note that when we restrict the above label network to have  one layer depth the projection is equivalent to the label mapping from previous work in Eq.~\ref{joint_form}.

\begin{figure*}
 	\centering
 	\hspace{5mm}\includegraphics[scale=0.55]{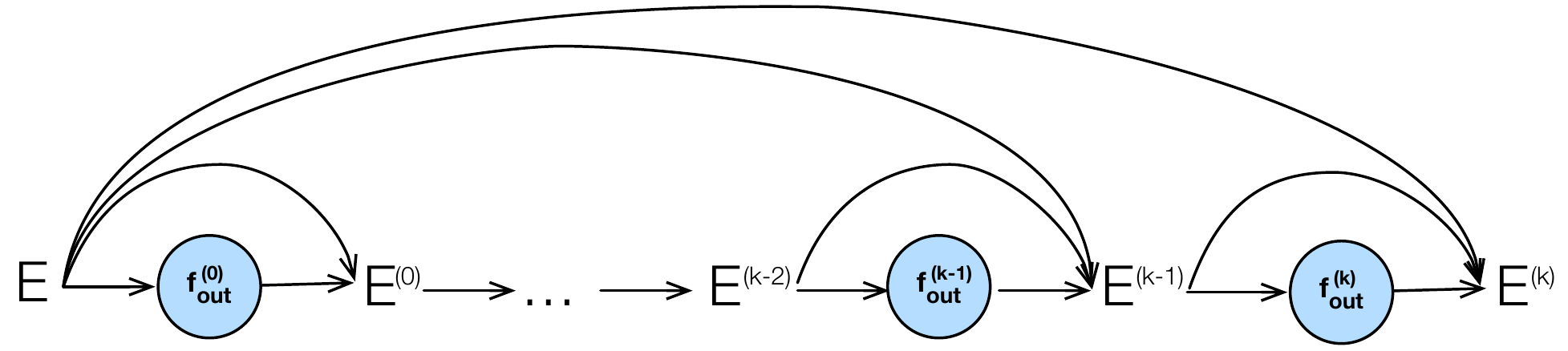}
 	\vspace{-2mm}
 	\caption{The proposed deep residual label network architecture for neural language generation. Straight lines represent the input to a function and curved lines represent shortcut or residual connections implying addition operations.}
 	\label{schema-mini}
\end{figure*}

\subsubsection{Preserving Information}
The multiple layers of projections in Eq.~\ref{kth_proj} force the relationship between word embeddings $\mathbf{E}$ and label embeddings $\mathbf{{E}^{(k)}}$ to be highly nonlinear. To preserve useful information from the original word embeddings and to facilitate the learning of the label network we add a skip connection directly to the input embedding. Optionally, for very deep label networks, we also add a residual connection to previous layers as in \citep{resnet15}. With these additions the projection at the $k_{th}$ layer becomes:
\begin{align} 
    \mathbf{{E}^{(k)}}    & = f^{(k)}_{out}(\mathbf{{E}^{(k-1)}}) + \mathbf{{E}^{(k-1)}} + \mathbf{E}
\end{align}

\subsubsection{Controlling Output Network Power}
We can characterize the power of the proposed output network in terms of its number of parameters $\Theta_{drill}$,  including the proposed label encoder and output classifier:
\begin{equation}
    C_{drill} \approx |\Theta_{drill}| = k \times (d \times d) + |\mathcal{V}|.
\end{equation}
By controlling the depth of the label encoder we can make the number of parameters equal to that of each other output network.  For weight tying, $C_{tied}$, this is $k=0$, for full linear weights, $C_{base}$, this is $k=\frac{|\mathcal{V}|}{d}$, for a bilinear mapping, $C_{bilinear}$, this is $k=1$, and for a dual nonlinear mapping, $C_{joint}$, this is $k=\frac{2d_j}{d}$.  Hence, the power of the output network can be adjusted freely depending on the task at hand within the full spectrum of options defined by Ineq.~\ref{capacity_ineq}. 

\subsubsection{Avoiding Overfitting}
The ability to increase power may be useful for high-resource data regimes, however it can lead to overfitting in the case we are in a low-resource data regime. To make sure that our network is robust to both data availability regimes, we choose to apply standard \cite{dropout15} or variational \cite{vardropout16} dropout in between each of the $k$ layers of the projection.  Assuming $\delta(\cdot)$ to be the dropout mask sampling function, the above goal is achieved by modifying the function $f^{(i)}_{out}(\cdot)$ at the $i_{th}$ layer from Eq~\ref{drill_proj} as follows:
 \begin{align}
    f^{\prime(i)}_{out}(\mathbf{E^{(i-1)}}) = \delta\big( f^{(i)}_{out}(\mathbf{E^{(i-1)}}) \big) \odot f^{(i)}_{out}(\mathbf{E^{(i-1)}}).
\end{align}   
In standard dropout, a new binary dropout mask is sampled every time the dropout function is called. This means that new dropout masks are sampled independently for each dimension of each different label representation. In contrast, variational dropout samples a binary dropout mask only once upon the first call and then repeatedly uses that locked dropout mask for all label representations within the forward and backward pass.

\subsection{Context Network} 
The context representation $\mathbf{h_t}$ in most language generation tasks is typically the output of a deep neural network, and thus it can capture, in principle, the nonlinear structure of the dual nonlinear mapping in Section~\ref{joint}. Eq.~\ref{joint} has an additional nonlinearity $g_{in}(\cdot)$ in order to allow the dimensionality of the joint space to be larger than that of the context encoder's output $\mathbf{h_t}$.  However, in our proposed model we increase the power of the output network by increasing the depth of the label encoder, keeping the size of the joint space fixed.  Thus, for our models, we make the simplifying assumption that there is no additional nonlinearity after the context encoder, setting $g_{in}(\cdot) = \mathbf{I}$.
 
\subsection{Training Considerations}
To perform maximum likelihood estimation of the model parameters, we use the negative log-likelihood of the data as our training objective.  This involves computing the conditional likelihood of predicting the next word, as explained above. The normalized exponential function we use for converting the network scores to probability estimates is the typical softmax activation function. 

In principle, our output layer parameterisation requires more computations than a typical softmax linear unit, with or without weight tying.  Hence, it tends to get slower as the depth of the label encoder or the size of the vocabulary increases. In case either of them becomes extremely large, we can resort to recent sampling-based or hierarchical softmax approximation methods such as the ones proposed by \citet{softapprox2015} and \citet{grave17}.
\footnote{Note that in practice, for our experiments with vocabularies up to 32K, we did not need to resort to a softmax approximation.}

\subsection{Relation to Previous Output Layer Forms}
Our output layer parameterisation has the same general form as the one with the dual nonlinear mapping in Eq.~\ref{general_joint_form}.
Hence, it preserves the property of being a generalization of output layers based on bilinear mapping and weight tying described in Section \ref{joint}.
The bilinear form in Eq.~\ref{eq:bilinear_soft} can be simply derived from the general form of Eq~\ref{drill_form} if we restrict the output mapping depth to be equal to one, set its bias equal to zero, and make the $\sigma(.)$ activation function linear, so we have $\mathbf{U^{(0)}} = \mathbf{W_l}$.
By further setting the matrix $\mathbf{U}$ to be the identity matrix, we can also derive the output layer form based on weight tying in Eq.~\ref{eq:tied}.

\begin{table*}[ht]%[!h]
	\small
	\centering
	\begin{tabular}{l|ccc}
		\toprule
		\bf Model & \bf \#Param & \bf Validation &  \bf Test \\
		\midrule
		\citet{mikolov2012context} -- RNN-LDA + KN-5 + cache & 9M$^\ddagger$ & - & 92.0 \\
		\citet{zaremba2014recurrent} -- LSTM & 20M & 86.2 & 82.7 \\
		\citet{gal2016theoretically} -- Variational LSTM (MC) & 20M & - & 78.6 \\
		\citet{kim2016character} -- CharCNN & 19M & - & 78.9 \\
		\citet{merity2016pointer} -- Pointer Sentinel-LSTM & 21M & 72.4 & 70.9 \\
		\citet{grave2016improving} -- LSTM + continuous cache pointer$^\dagger$ & - & - & 72.1 \\
		\citet{inan2016tying} -- Tied Variational LSTM + augmented loss & 24M & 75.7 & 73.2 \\
		\citet{zilly2016recurrent} -- Variational RHN & 23M & 67.9 & 65.4 \\
		\citet{zoph2016neural} -- NAS Cell & 25M & - & 64.0 \\
		\citet{melis2017state} -- 2-layer skip connection LSTM & 24M & 60.9 & 58.3 \\
		\midrule
		\citet{merity2017regularizing} -- AWD-LSTM w/o finetune & 24M & 60.7 & 58.8 \\%
		\citet{merity2017regularizing} -- AWD-LSTM & 24M & 60.0 & 57.3 \\
		Ours -- AWD-LSTM-DRILL w/o finetune & 24M & 59.6 & 57.0 \\% 59.64 & 57.05
		Ours -- AWD-LSTM-DRILL  & 24M & \textbf{58.2} & \textbf{55.7} \\%58.26/55.74

		\midrule
	
		\citet{merity2017regularizing} -- AWD-LSTM + continuous cache pointer$^\dagger$ & 24M & 53.9 & 52.8 \\
		\citet{krause2017dynamic} -- AWD-LSTM + dynamic evaluation$^\dagger$ & 24M & 51.6 & 51.1 \\
		Ours -- AWD-LSTM-DRILL + dynamic evaluation$^\dagger$ & 24M & \textbf{49.5} & \textbf{49.4}  \\

		\midrule \midrule
		\citet{mos2018} -- AWD-LSTM-MoS & 22M & {56.54} & {54.44} \\
		\citet{mos2018} -- AWD-LSTM-MoS + dynamic evaluation$^\dagger$ & 22M & {48.33} & {47.69} \\
		\bottomrule
	\end{tabular}
	\vspace{-2mm}
	\caption{\small
		Model perplexity with a single softmax (upper part) and multiple softmaxes (lower part) on validation and test sets on Penn Treebank. Baseline results are obtained from \citet{merity2017regularizing} and \citet{krause2017dynamic}. $\dagger$ indicates the use of dynamic evaluation.
	}
		\vspace{-4mm}
	\label{table:PTB}
\end{table*}

\section{Experiments}
\label{eval}

We evaluate on three language generation tasks.  The first two tasks are standard language modeling tasks, i.e.~predicting the next word given the sequence of previous words.  The third task is a conditional language modeling task, namely neural machine translation, i.e.~predicting the next word in the target language given the source sentence and the previous words in the translation.  To demonstrate the generality of the proposed output mapping we incorporate it in three different neural architectures which are considered state-of-the-art for their corresponding tasks.

\begin{table*}[t]%[!h]
	\small
	\centering
	\begin{tabular}{l|ccc}
		\toprule
		\bf Model & \bf \#Param & \bf Validation &  \bf Test \\
		\midrule
		\citet{inan2016tying} -- Variational LSTM  + augmented loss & 28M & 91.5 & 87.0 \\
		\citet{grave2016improving} -- LSTM + continuous cache pointer$^\dagger$ & - & - & 68.9 \\
		\citet{melis2017state} -- 2-layer skip connection LSTM & 24M & 69.1 & 65.9 \\
		\midrule
		\citet{merity2017regularizing} -- AWD-LSTM w/o finetune & 33M & 69.1 & 66.0 \\
		\citet{merity2017regularizing} -- AWD-LSTM & 33M & 68.6 & 65.8 \\
		Ours -- AWD-LSTM-DRILL w/o finetune & 34M & {65.7}  & {62.8}  \\ %65.70/62.81
		Ours -- AWD-LSTM-DRILL  & 34M & \textbf{64.9}  & \textbf{61.9}  \\ %64.93/61.98
		\midrule
		\citet{merity2017regularizing} -- AWD-LSTM + continuous cache pointer $^\dagger$& 33M & 53.8 & 52.0 \\
		\citet{krause2017dynamic} -- AWD-LSTM + dynamic evaluation$^\dagger$ & 33M & 46.4 & 44.3 \\
		Ours -- AWD-LSTM-DRILL + dynamic evaluation$^\dagger$ & 34M & \textbf{43.9} &  \textbf{42.0}  \\
		\midrule \midrule
		\citet{mos2018} -- AWD-LSTM-MoS & 35M & {63.88} & {61.45} \\
		\citet{mos2018} -- AWD-LSTM-MoS + dynamical evaluation$^\dagger$ & 35M & {42.41} & {40.68} \\
		\bottomrule
	\end{tabular}
	\vspace{-1mm}
	\caption{\small 
		Model perplexity with a single softmax (upper part) and multiple softmaxes (lower part) on validation and test sets on WikiText-2. Baseline results are obtained from \citet{merity2017regularizing} and \citet{krause2017dynamic}. $\dagger$ indicates the use of dynamic evaluation.
	}
	\vspace{-4mm}
	\label{table:WT2}
\end{table*}
 
\subsection{Language Modeling}
\label{sec:lm}

\textbf{Datasets and Metrics}. Following previous work in language modeling  \cite{mos2018,krause2017dynamic,merity2017regularizing,melis2017state}, we evaluate the proposed model in terms of  perplexity on two widely used language modeling datasets, namely Penn Treebank \cite{mikolov2010} and WikiText-2  \cite{merity2017regularizing} which have vocabularies of 10,000 and 33,278 words, respectively.  For fair comparison, we use the same regularization and optimization techniques with  \citet{merity2017regularizing}. 

\textbf{Model Configuration}.
To compare with the state-of-the-art we use the proposed output layer within the best architecture by \citet{merity2017regularizing}, which is a highly regularized 3-layer LSTM with 400-dimensional embeddings and 1150-dimensional hidden states, noted as AWD-LSTM. Our hyper-parameters were optimized based on validation perplexity, as follows:  4-layer label encoder depth, 400-dimensional label embeddings,  0.6 dropout rate,  residual connection to $\mathbf{E}$, uniform weight initialization in the interval $[-0.1, 0.1]$, for both datasets, and, furthermore, \textit{sigmoid} activation and \textit{variational} dropout for PennTreebank, as well as \textit{relu} activation and \textit{standard} dropout for Wikitext-2. The rest of the hyper-parameters were set to the optimal ones found for each dataset by \citet{merity2017regularizing}. 

For the implementation of the AWD-LSTM we used the language modeling toolkit in  Pytorch provided by \citet{merity2017regularizing},\footnote{\url{http://github.com/salesforce/awd-lstm-lm}} and for the dynamic evaluation the code in Pytorch provided by \citet{krause2017dynamic}.\footnote{\url{http://github.com/benkrause/dynamic-evaluation}} 

\subsubsection{Results}
\label{lm_results}

The results in terms of perplexity for our models, denoted by DRILL, and several competitive baselines, are displayed in Table \ref{table:PTB} for PennTreebank and Table \ref{table:WT2} for Wikitext-2.
For the single-softmax models (above the double lines), for both datasets, our models improve over the state-of-the-art by +1.6 perplexity on PennTreebank and by +3.9 perplexity on Wikitext-2. Moreover, when our model is combined with the dynamic evaluation approach proposed by \citet{krause2017dynamic}, it improves even more over these models by +1.7 on PennTreebank and by +2.3 on Wikitext-2.

In contrast to other more complicated previous models, our model uses a standard LSTM architecture, following the work of  \citet{merity2017regularizing,melis2017state}. For instance, \citet{zilly2016recurrent} uses of a recurrent highway network which is an extension of an LSTM to allow multiple hidden state updates per time step,  \citet{zoph2016neural} uses reinforcement learning to generate an RNN cell which is even more complicated than an LSTM cell, and \citet{merity2016pointer} makes use of a probabilistic mixture model which combines a typical language model with a pointer network  which reproduces words from the recent context.

Interestingly, our model also significantly reduces the performance gap against multiple softmax models.  In particular, when our finetuned model is compared to the corresponding mixture-of-softmaxes (MoS) model, which makes use of 15 softmaxes in the classifier, it reduces the difference against AWD-LSTM from 2.8 to 1.2 points on PennTreebank and from 4.3 to 0.4 points  on WikiText-2. When our model is compared to MoS with dynamic evaluation, the difference is reduced from 3.4 points to 1.7 points on PennTreebank and from 3.6 to 1.3 on WikiText-2. Note that the rank of the log-probability matrix for MoS on PennTreebank is 9,981, while for AWD-LSTM and our model the rank is only 400. This observation questions the high-rank hypothesis of MoS, which states that the log-probability matrix has to be high rank to better capture language. Our results suggest that the log-probability matrix does not need to be high rank if the classifier is better at capturing the output space structure.
 
Furthermore, as shown in Table~\ref{table:speed}, the MoS model is far slower than AWD-LSTM, even for these small datasets and reduced dimensionality settings,\footnote{
Note that even though the MoS models have a comparable number of parameters to the other models, they use smaller values for several crucial hyper-parameters, such as word embedding size, hidden state size and batch size, likely to make the training speed more manageable and avoid overfitting.  
}\begin{table}[htp]%[!h]
	\centering
	\small
	 {\def\arraystretch{1.1}\tabcolsep=3.5pt\begin{tabular}{l|ll}
		\toprule
		\bf Model     & \bf PennTreebank &  \bf Wikitext-2 \\
		\midrule
		AWD-LSTM       & 47 sec ($1.0\times$) & 89 sec ($1.0\times$) \\
		AWD-LSTM-DRILL & 53 sec ($1.1\times$) & 106 sec ($1.2\times$) \\
		AWD-LSTM-MoS   & 139 sec ($3.0\times$) & 862 sec ($9.7\times$)\\
 		\bottomrule
	\end{tabular}} 
	\vspace{-1mm}
	\caption{\small 
		Average time taken per epoch on the two datasets: PennTreebank ($|\mathcal{V}|\approx20K$) and Wikitext-2 ($|\mathcal{V}|\approx33K$). 
	} 
	\vspace{-6mm}
	\label{table:speed}
\end{table}
whereas adding our label encoder to AWD-LSTM results in only a small speed difference. In particular, on PennTreebank the MoS model takes about 139 seconds per epoch while AWD-LSTM about 47 seconds per epoch, which makes it slower by a factor of $3.0\times$, whereas our model is only $1.1\times$ slower than this baseline.  On Wikitext-2, the differences are even more pronounced due to the larger size of the vocabulary.  The MoS model takes about 862 seconds per epoch while AWD-LSTM takes about 89 seconds per epoch, which makes it slower by a factor of $9.7\times$, whereas our model with 4-layers is only $1.2\times$ slower than the baseline. We attempted to combine our label encoder with the MoS model, but its training speed exceeded our computation budget. 

Overall, these results demonstrate that the proposed deep residual output mapping improves significantly the state-of-the-art single-softmax neural architecture for language modeling, namely AWD-LSTM, without hurting its efficiency.  Hence, it could be a useful and practical addition to other existing architectures. In addition, our model remains competitive against models based on multiple softmaxes and could be combined with them in the future, since our work is orthogonal to using multiple softmaxes. To demonstrate that our model is also applicable to larger datasets as well, in Section~\ref{nmt} below we apply our method to neural machine translation. But before moving to that experiment, we first perform an ablation analysis of these results.

\begin{table}[t]%[!h]
	\small
	\centering
	{\def\arraystretch{1.1}\tabcolsep=3.5pt\begin{tabular}{l|ccc}
		\toprule
		\bf Output Layer & \bf \#Param & \bf Validation &  \bf Test \\
		\midrule
        Full softmax                    & 43.8M & 69.9 & 66.8 \\
        Weight tying [PW17]             & 24.2M & 60.0 & 57.3 \\
        Bilinear map. [G18]             & 24.3M & 60.7 & 58.5  \\  
        Dual nonlinear map. [PH18]      & 24.5M & 58.8 & 56.4  \\ \hline
        DRILL 1-layer                    & 24.3M & 58.8 & 56.2 \\ 
        DRILL 2-layers                   & 24.5M & 58.7 & 56.0\\ 
        DRILL 3-layers                   & 24.7M & 58.5 & 55.9\\ 
        DRILL 4-layers                   & 24.8M & \textbf{58.2} & \textbf{55.7} \\ 
        \hspace{2mm}  + residuals between layers   & 24.8M & 59.6 & 57.5 \\ 
        \hspace{2mm}  - no variational dropout        & 24.8M & 63.4 & 60.7\\ 
		\bottomrule
	\end{tabular}}
	\vspace{-2mm}
	\caption{\small 
	Ablation results and comparison with previous output layers when using AWD-LSTM \cite{merity2017regularizing} as an encoder network on PennTreebank.
	}
	\label{table:ablation}
	\vspace{-4mm}
\end{table}

\subsubsection{Ablation Analysis}

To give further insights into the source of the improvement from our output layer parameterisation, in Table~\ref{table:ablation} we compare its ablated variants with previous output layer parameterisations.  Each alternative is combined with the state-of-the-art encoder network AWD-LSTM \citep{merity2017regularizing}. We observe that full softmax produces the highest perplexity scores, despite having almost 20M parameters more than the other models. This shows that the power of the output layer or classifier, as measured by number of parameters, is not indicative of generalization ability. 

The output layer with weight tying \cite{press17}, noted [PW17], has lower perplexity than the full softmax by 9.5 points. The bilinear mapping \cite{gulordava18}, noted [G18], has lower perplexity than the full softmax by 8.3 points, but it is still higher than weight tying by 1.2 points.  The dual nonlinear mapping \cite{pappas18}, noted [PH18], has even lower perplexity compared to the full softmax by 10.4 points, and has lower perplexity than weight tying by 0.9 points.\footnote{For fair comparison, we also used dropout and residual connections to $E$ and $h_t$ when they lead to better validation performance.} 
DRILL with only 1-layer depth is slightly better than [PH18], and with 2-layers depth outperforms all previous output mappings, improving over full softmax by 10.8 points, weight tying by 1.3 points, and dual non-linear mapping by 0.4 points. Increasing the depth even more provides further improvements of up to 0.3 points. This shows the benefits of learning deep output label mappings, as opposed to shallower ones. Lastly, DRILL with residual connections between layers  has an increase of 1.8 perplexity points, likely because of an effective reduction in depth, and not using variational dropout has a significant increase in perplexity of namely 5 points, which highlights the importance of regularization between layers for this task.

\begin{figure}[htp]
\vspace{-3mm}
 	\centering
 	\hspace{-4mm}\includegraphics[scale=0.5]{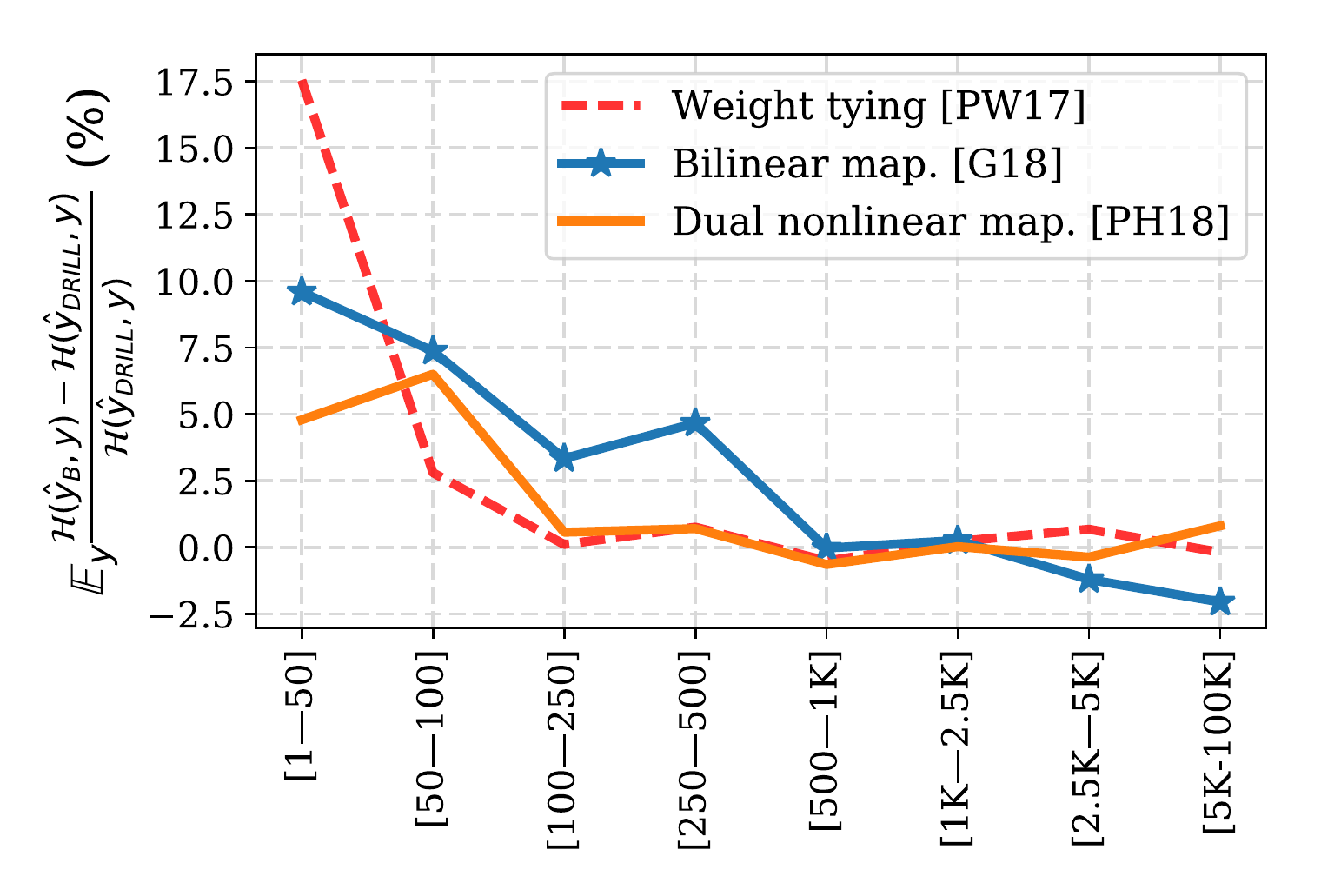}
   \vspace{-7mm}
 	\caption{Mean relative cross-entropy loss difference (\%) between each baseline output layer (B) and our output layer (DRILL) computed over different word frequency intervals on PennTreebank.}
 	\label{lossdiff}
	\vspace{-4mm}
\end{figure}
To verify the hypothesis that our output layer facilitates information transfer across words, we also  analyzed the loss for words in different frequency bands, created by computing statistics on the training set. Figure \ref{lossdiff} displays the mean relative cross-entropy difference (\%) between our output layer and the previous output layers for the different word frequency bands on the test set of PennTreebank.  Overall, the graph shows that most of the improvements in perplexity between 5\% to 17.5\% brought by DRILL against baselines comes from predicting more accurately the words in lower word frequency bands (1 to 100 occurrences). The results are consistent with Table \ref{table:ablation}, since the second best output layer is the one with the bilinear mapping followed by the bilinear mapping and weight tying baselines. One exception occurs in the highest frequency band, where DRILL has 2.5\% higher perplexity than the bilinear mapping, but this difference is less significant because it is computed based on 16 unique words as opposed to the lowest frequency band which corresponds to 4116 unique words. 
These results validate our hypothesis that learning a deeper label encoder leads to better transfer of learned information across labels.  More specifically, because low frequency words lack data to individually learn the complex structure of the output space, transfer of learned information from other words is crucial to improving performance, whereas this is not the case for higher frequency words.
This analysis suggests that our model could also be useful for zero-resource scenarios, where labels need to be predicted without any training data, similarly to other joint input-output space models.

\subsection{Neural Machine Translation}
\label{nmt}

\textbf{Dataset and Metrics}. 
Following previous work in neural machine translation \cite{NIPS2017_7181}, we train on the WMT 2014 English-German dataset with $\sim$4.5M sentence pairs, using the Newstest2013 set for validation and the Newstest2014 set for testing. We pre-process the texts using the BPE algorithm  \cite{sennrich15} with 32K  operations. Following the standard evaluation practices in the field \cite{bojar-EtAl:2017:WMT1}, the translation quality is measured using BLEU score  \citep{papineni-EtAl:2002:ACL} on \emph{tokenized} text.

\textbf{Model configuration}.
We compare against the state-of-the-art Transformer (base) architecture from  \citet{NIPS2017_7181} with a 6-layer encoder and decoder depth, 512-dimensional word embeddings, 2048-dimensional hidden feed-forward states and 8 heads.\footnote{We chose the base model because it can be trained much faster than the big model (12 hours vs 3.5 days), for efficiency reasons.} Our hyper-parameters were optimized based on validation accuracy, as follows: 2-layer label encoder depth, 512-dimensional label embeddings, 0.0 dropout rate, \textit{sigmoid} activation function, residual connection to $\mathbf{E}$, and uniform weight initialization in $[-0.1, 0.1]$. The rest of the hyper-parameters were set to the optimal ones in \citep{NIPS2017_7181}, except that we did not perform model averaging over last 5  for Transformer (base) model. To ensure fair comparison, we trained the Transformer (base) from scratch for the same number of training steps as ours, namely 350K, and thereby reproduced about the same score as in \citep{NIPS2017_7181} with a slight difference of +0.1 point. 
For the implementation of the Transformer, we used OpenNMT %the opennmt library written in pytorch  
\citep{opennmt}.\footnote{\url{http://github.com/OpenNMT/OpenNMT-py}}

\subsection{Results}

The results displayed in Table~\ref{table:nmt_results} show that our model, namely Transformer-DRILL (base) with 79.9M parameters, outperforms the Transformer (base) model with 79.4M parameters by 0.8 points, and is only 0.3 points behind the Transformer (big) model which has 240 parameters due to its increased dimensionality. This result almost matches the single-model state-of-the-art, without
resorting to very high capacity encoders or model averaging over different epochs. Transformer-DRILL also outperforms by 0.6 points our implementation of Transformer (base) model combined with the dual nonlinear mapping by \citet{pappas18}, highlighting once more the importance of deeper label mappings. Note that our improvement is noticeable even when the vocabulary is based on sub-word units \cite{sennrich15}, instead of regular word units as in Section \ref{lm_results}. 

Lastly, our model even surpasses the performance of some ensemble models such as GNMT + RL and ConvS2S. The RNMT+ model is marginally better than Transformer (big) even though it has two layers deeper decoder and more powerful layers, namely bidirectional LSTMs instead of self-attention. RNMT+ cascaded and multicol are ensemble architectures which combine LSTMS with self-attention in different ways and increase the overall model complexity even more while providing marginal gains over simpler architectures. Combining our output layer with Transformer (big) should, in principle, make this difference even smaller.

\section{Other Related Work}
Several studies focus on learning the structure of the output space from texts for zero-shot classification \cite{dauphin14,nam16,rios-kavuluru-2018-shot,pappas19b} and structured prediction \cite{Srikumar14,dyer2015,YehWKW17}. Fewer such studies exist for neural language generation, for instance the ones described in Section \ref{background}. Their mappings can increase the power of the classifier by controlling its dimensionality or rank, but unlike ours, they have limited expressivity and a tendency to overfit. \citet{mos2018} showed that the softmax layer which is low-rank creates a `bottleneck' problem, i.e.~limits model expressivity, and increased the classifier rank by using a mixture of softmaxes. \citet{takase18} improved MoS by computing the mixture based on the last and the middle recurrent layers. Two alternative ways to increase the classifier rank are obtained by  multiplying the softmax with a non-parametric sigmoid function \cite{NIPS2018_7312}, and by learning parametric monotonic functions on top of the logits \cite{ganea}. Both of these methods have close to or higher perplexity than ours without using MoS, even though we keep the rank or power of the classifier the same. Instead, we specifically increase the power of the output label encoder, and the obtained results suggest that the classifier does not necessarily need to be high-rank to better capture language. 

\begin{table}
\begin{tabular}{l  c  }
\toprule
\textbf{Model} & \textbf{BLEU}  \\
\hline
Bidirectional GRU \citep{sennrich15} & 22.8   \\
ByteNet \citep{NalBytenet2017} & 23.7   \\
GNMT + RL \citep{TACL1081} & 24.6   \\
ConvS2S \citep{JonasFaceNet2017} & 25.1 \\
MoE \citep{shazeer2017outrageously} & 26.0   \\
GNMT + RL Ensemble \citep{TACL1081} & 26.3 \\
ConvS2S Ensemble \citep{JonasFaceNet2017} & 26.3  \\
\specialrule{1pt}{-1pt}{0pt}
\rule{0pt}{2.2ex}Transformer (base) \citep{NIPS2017_7181} & 27.3  \\
\rule{0pt}{2.2ex}Transformer-Dual (base) [PH18] & {27.5} \\ 
\rule{0pt}{2.2ex}Ours -- Transformer-DRILL (base) & \textbf{28.1} \\ \hline \hline
\rule{0pt}{2.2ex}Transformer (big) \citep{NIPS2017_7181} & 28.4  \\ 
\rule{0pt}{2.2ex} RNMT+ \citep{chen2018} & 28.5  \\ 
\rule{0pt}{2.2ex} RNMT+ cascaded \citep{chen2018} & 28.6  \\
\rule{0pt}{2.2ex} RNMT+ multicol \citep{chen2018} & 28.8  \\
\bottomrule
\vspace{-7mm}
\end{tabular}
\caption{Translation results in terms of BLEU on English to German with a 32K BPE vocabulary.}
\label{table:nmt_results}
\vspace{-5mm}
\end{table}

\section{Conclusion}

Typical log-linear classifiers for neural language modeling tasks can be significantly improved by learning a deep residual output label encoding, regardless of the input encoding architecture.  Deeper representations of the output structure lead to better transfer across the output labels, especially the low-resource ones. The results on three tasks show that the proposed output layer parameterisation can match or improve state-of-the-art context encoding architectures and outperform previous output layer parameterisations based on a joint input-output space, while preserving their basic principles and generality. Our findings should apply on other conditional neural language modeling tasks, such as image captioning and summarization. As future work, it would be interesting to learn from  more elaborate descriptions or contextualized representations of the output labels and investigate their transferability in different tasks.

\section*{Acknowledgements}
This work was supported by the European Union through SUMMA project (n.~688139) and the Swiss National Science Foundation within INTERPID project (FNS-30106).  

\bibliography{references}
\bibliographystyle{icml2019}

\end{document}